\documentclass[letterpaper, 10 pt, conference]{ieeeconf}  % Comment this line out if you need a4paper

\usepackage[top=54pt,bottom=54pt,left=54pt,right=54pt]{geometry}

\usepackage{amssymb, amsmath, amsthm}

\usepackage{cite}
\usepackage{graphicx}
\usepackage{siunitx}

\usepackage{blindtext}
\usepackage{algorithm}
\usepackage{algpseudocode}
\usepackage{verbatim}
\usepackage{soul}	%\hl{}

\usepackage{makecell}

\usepackage{hyperref}

\usepackage{color}

\newtheorem{definition}{Definition}
\newtheorem{proposition}{Proposition}

\theoremstyle{definition}

\newcommand{\calA}{{\cal A}}
\newcommand{\calC}{{\cal C}}
\newcommand{\calV}{{\cal V}}

\def\d{\mathrm{d}}
\renewcommand{\vec}[1]{\mathbf{#1}}

\IEEEoverridecommandlockouts                              % This command is only needed if 
\title{\LARGE \bf
Robotic Manipulation of a Rotating Chain with Bottom End Fixed
\thanks{The authors would like to thank Dr. Hung Pham from Eureka Robotics for insightful discussions that helped establish the foundational understanding necessary for this work.}}

\author{Qi Jing Chen$^{1}$, Shilin Shan$^{1}$, and Quang-Cuong Pham$^{2}$% <-this % stops a space
\thanks{*Github repo: \url{https://github.com/qj25/rotatingchain_wbef.git} (Videos: \url{https://youtu.be/Hhag8Ea5x30})}% <-this % stops a space
\thanks{$^{1}$Nanyang Technological University, School of Mechanical and Aerospace Engineering,
$^{2}$Eureka Robotics, Singapore
	}%
}

\newcommand{\cmmnt}[1]{\ignorespaces}
\begin{document}

\maketitle
\thispagestyle{empty}
\pagestyle{empty}

%%%%%%%%%%%%%%%%%%%%%%%%%%%%%%%%%%%%%%%%%%%%%%%%%%%%%%%%%%%%%%%%%%%%%%%%%%%%%%%%
\begin{abstract}
This paper studies the problem of using a robot arm to manipulate a uniformly rotating chain with its bottom end fixed. Existing studies have investigated ideal rotational shapes for practical applications, yet they do not discuss how these shapes can be consistently achieved through manipulation planning. Our work presents a manipulation strategy for stable and consistent shape transitions. We find that the configuration space of such a chain is homeomorphic to a three-dimensional cube. Using this property, we suggest a strategy to manipulate the chain into different configurations, specifically from one rotation mode to another, while taking stability and feasibility into consideration. We demonstrate the effectiveness of our strategy in physical experiments by successfully transitioning from rest to the first two rotation modes. The concepts explored in our work have critical applications in ensuring safety and efficiency of drill string and yarn spinning operations.
\end{abstract}

\section{Introduction}
\label{sec:intro}

Rotating chains are fascinating objects that have been studied from both the mathematical \cite{kolodner1955heavy,stuart2006steadily,toland1979stability,caughey1958whirling,caughey1969whirling,wu1972whirling} and engineering \cite{pham2017robotic} perspectives. Mathematical analysis has shown that a chain rotated from the top end and with a free bottom end can experience several modes of rotation, characterized by the number of times the chain intersects its rotation axis~\cite{kolodner1955heavy}. Further research into its configuration space established a strategy for robotic manipulation of the chain to transition between the different rotation modes~\cite{pham2017robotic}. However, rotating chains with free bottom ends have relatively few practical applications. Indeed, real-world instances of long, flexible, rotating objects (of which rotating chains are the models) usually have both ends fixed: such are the cases of drill strings or spun yarns. The reasons contributing to drill string fatigue and yarn thread breakage in the spinning process are well-documented. While existing research explores optimal rotational shapes for both operations, a significant gap remains in understanding how manipulators can achieve these shapes in a controlled manner. Uncontrolled manipulation can result in unstable rotations causing unnecessary wear in drill strings, or generate wrong mode shapes that induce high tension and cause breakage in yarn threads.

This paper investigates the problem of planning stable and controlled manipulation for rotating chains with fixed bottom ends. By adapting the arguments of~\cite{pham2017robotic} to this case, we establish that the configuration space of the rotating chain with bottom end fixed is homeomorphic to a three-dimensional cube with fixed height. We then develop a methodology to address both dynamic and kinematic constraints of the manipulator within the stable configuration space. We verify our results with physical experiments by transiting from the first to second rotation mode in a smooth and precise manner. Our work can prevent issues such as mechanical fatigue and operational hazards in drill strings by optimizing rotation speed and amplitude to ensure stability and safety. Additionally, by focusing on avoiding high-tension shapes during yarn spinning, we propose strategies to enhance the efficiency of spinning operations and reduce the risk of breakage (see Section~\ref{sec:frc-appl}). This paper aims to bridge the gap between theoretical models and practical applications, offering solutions that improve the reliability and efficiency of handling rotating chains in industrial contexts.

\begin{figure}[tp]
	\centering
	\textbf{a}\\[0.02in]
	\includegraphics[width=0.47\textwidth]{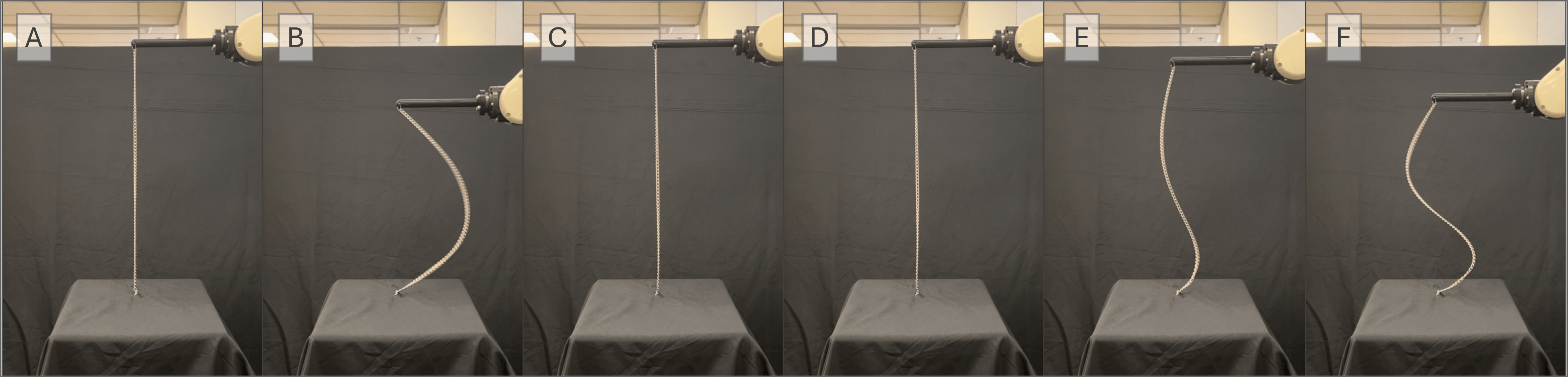}\\
	\textbf{b}\\
	\includegraphics[width=0.47\textwidth]{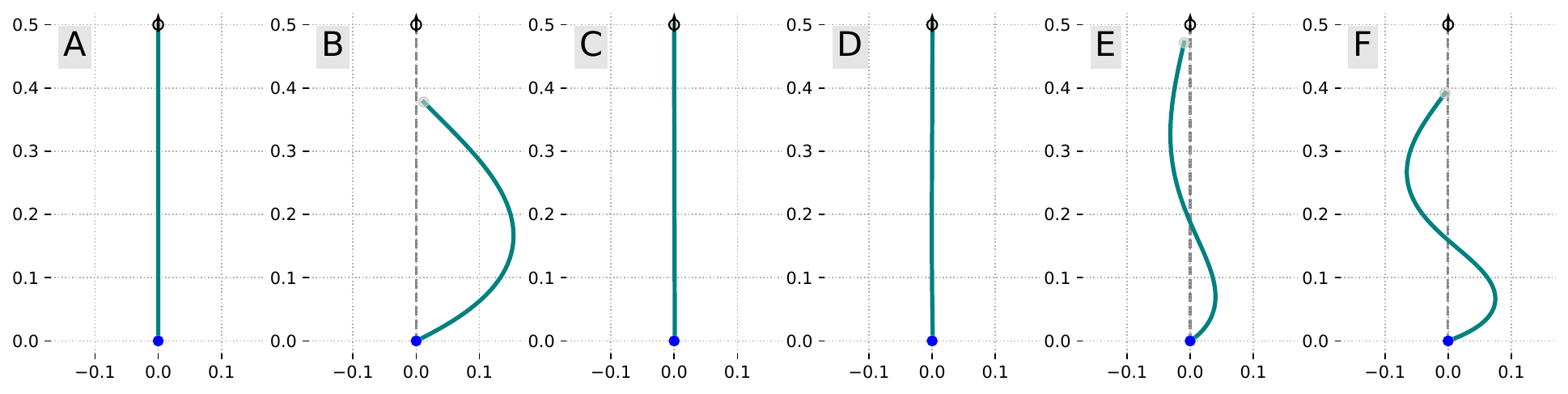}
	\caption{\textbf{a}: Robotic manipulation of a rotating chain with bottom end fixed along a vertical axis and top end attached to a robot end-effector rotating along the vertical axis. Pictured are real configurations experienced when following the manipulation strategy (A to F) proposed in Section~\ref{sec:experiment} for transition to the first two rotation. A video of the physical experiments can be found at \url{https://youtu.be/Hhag8Ea5x30}. \textbf{b}: Corresponding simulated configurations obtained from forward kinematics of the rotating chain. Although they have similar shapes, A, C, and D are not of the same configuration due to differences in their angular velocities.}
	\label{fig:realshapes}
	\vspace{-0.5cm}
\end{figure}

%\begin{figure}[t]
%	\centering
%	\includegraphics[width=0.45\textwidth]{pics/env_start_success5.png}
%	\caption{Simulation environment with a wire and two obstacles in MuJoCo for the flinging task. The start (a) and a successful end (d) configuration are shown.}
%	\label{fig:envstartsuccess}
%\end{figure}
\subsection*{Contribution and organization of the paper}
The contribution of our paper is threefold. First, we show that the configuration space of a rotating chain with bottom end fixed is homeomorphic to a three-dimensional cube with fixed height (Section~\ref{sec:configuration_space}). 

Second, we present a manipulation strategy for stable shape and mode transitions of the rotating chain (Section~\ref{sec:manipulation}). This is achieved by defining a subspace within the stable configuration space that respects the robot manipulator's dynamic and kinematic constraints.

Third, our work offers practical solutions for industrial applications involving rotating chains with fixed bottom ends, such as drill strings and spun yarns (Section~\ref{sec:implappl}).

% This includes the case of drill strings where operational stability and safety can be improved by managing rotation speed and amplitude, and the yarn spinning process where reducing breakage and increasing efficiency can be achieved by avoiding high-tension configurations.

%%%%%%%%%%%%%%%%%%%%%%%%%%%%%%%%%%%%%%%%%%%%%%%%%%%%%%%%%%%%%%%%%%%%%%%%%%%%%%%%

\section{Related works}
\label{sec:related-works}

Studies related to the rotating chain can be found in numerous fields including robotic manipulation of deformable linear objects, mathematical analysis of a rotating chain, and research on drill strings and yarn spinning. The following section will review the relevant literature in these fields.

\subsection{Robotic manipulation of deformable linear objects}
\label{sec:robotic-manip}
Deformable linear objects (DLOs) tend to have complex dynamics and studies on their manipulation commonly face the problem of underactuation. To circumvent this issue, topology was employed in the context of knotting~\cite{saha2007manipulation} and untangling~\cite{huang2023untangling}. The subject of homeomorphism~\cite{bretl2014quasi} and path-connectedness in topological spaces can play an important role in simplifying the configuration space of a Kirchhoff elastic rod, the latter proving useful for work in robotic DLO manipulation~\cite{borum2015free}.

In recent years, machine learning is a tool more commonly employed. The complex dynamics of a DLO can be handled by using algorithms that train models for perception or manipulation. Such techniques are able to take in whole scene states (e.g., through an image), and output a desirable robot action for both quasistatic~\cite{han2017model} and dynamic~\cite{zhang2021robots,chi2022iterative} manipulations.

Our work is framed within the analytical perspective found under the topic of kinematics. We study the characteristics of the configuration space of the rotating chain with bottom end fixed by examining its dynamic model, thus drawing conclusions on the space useful for changing shapes and modes through robotic manipulation.

\subsection{Mathematical analysis of the rotating chain}
\label{sec:applied-mathematics}
The study of rotating chains began with studies on the relation of angular speeds with rotation modes and findings that uniform rotations require a minimum angular speed more than the critical speed~\cite{kolodner1955heavy,toland1979stability,stuart2006steadily}. This research domain was later divided into two distinct areas: freely rotating chains~\cite{caughey1958whirling,wu1972whirling} and chains subjected to constant axial tension~\cite{caughey1969whirling}. Although these works contain in-depth analyses of the shapes formed by rotating chains, they lack comprehensive investigations into how the chains can be manipulated to form these shapes -- a crucial component for practical applications.

This problem was partially addressed through the investigation of the configuration space of a freely rotating chain~\cite{pham2017robotic}, which facilitated rotation mode transitions via robotic manipulation. Our work will study the case of a rotating chain with bottom end fixed, drawing insightful conclusions about its configuration space and considering feasibility within this newly-formed space to ensure workable physical experiments.

\subsection{Industrial applications}
\label{sec:frc-appl}
\subsubsection{Drill string operation}
Drill strings are a vital part of drill rigs which are used for gas and oil mining operations. To better understand them, an analytical study was conducted which numerically derived graphical solutions for their natural frequencies~\cite{bailey1960analytical}. Research on drill string fatigue has identified its primary causes as high-frequency oscillations~\cite{wada2018longitudinal}, and collisions with the borehole walls~\cite{liao2012parametric,dong2016review}. While the current work provides valuable insights, our research aims to enhance its applicability and usefulness by identifying the conditions under which a drill string configuration can effectively minimize fatigue, and how manipulation planning can avoid conditions which do the opposite.

%The techniques suggested in our study can improve drill string performance by functioning as a guide on how to minimize both these scenarios. Generally, we find that to reduce drill string wear, the rotation speed of the manipulated end should be limited to avoid high-frequency oscillations. In addition, a smaller rotation amplitude is preferred to reduce collision of the drill string with the borehole. To prevent sudden, uncontrolled collisions of the drill string with the borehole, unstable rotations should be avoided. 

\subsubsection{Yarn spinning}
Yarn balloons, loops formed by rotating yarn, are crucial for maintaining tension during the yarn spinning process. They form shapes similar to that of a rotating chain with bottom end fixed. Dynamic analysis of the yarn balloons was conducted to explore their regions of instability~\cite{batra1989integrated}. To predict the effects of air-drag, a mathematical model of the yarn balloon was established~\cite{tang2007modelling}. Studies on the relationship of tension in the spinning yarn with its control and shape parameters experimentally demonstrated the force differences in the yarn between two different rotation modes~\cite{wang2024experimental}. The work is commendable for its recommendation of yarn balloon shapes which prevent breakage caused by tension and wear. However, it fails to address the challenge of planning a stable control path to consistently achieve the desired shapes, a problem our research seeks to resolve. Using our theory of the rotating chain with bottom end fixed, Section~\ref{sec:implappl} further discusses how we can reduce fatigue in drill strings, and yarn breakage.

%%%%%%%%%%%%%%%%%%%%%%%%%%%%%%%%%%%%%%%%%%%%%%%%%%%%%%%%%%%%%%%%%%%%%%%%%%%%%%%%
\section{Problem Definition}
\label{sec:background}
This section will define the manipulation problem faced by the rotating chain with bottom end fixed in its general form. Subsequently, we will explore how this form can be modified for direct application to its various industrial settings.

\subsection{Equations of motion of the rotating chain}
Recall the equations of motion which govern the rotating chain as defined in~\cite{kolodner1955heavy}, where the chain has length $L$ and linear density $\mu$. This chain rotates about a vertical \textit{z}-axis. The top end of the chain is kept from the \textit{z}-axis at a distance of $r$ defined as the attachment radius, while the bottom end is fixed along the \textit{z}-axis.

\begin{figure}[htp]
	\centering
	\vspace{-0.2cm}
	\includegraphics[width=0.26\textwidth,trim={0 0 0 0},clip]{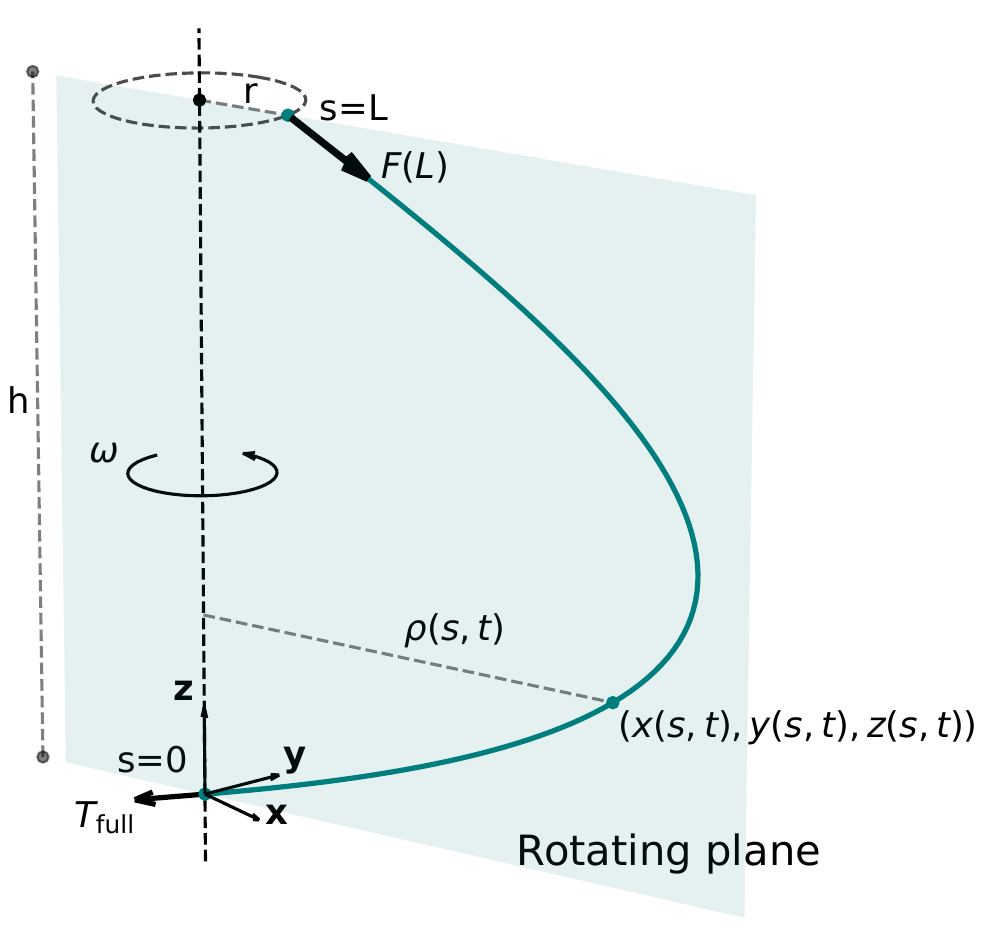}
	\vspace{-0.3cm}
	\caption{A chain, with its bottom end fixed, rotating about a fixed vertical axis. A 3D curve parameterized by $s$ is shown at a time instant $t$: $s=0$ at the bottom end, $s=L$ at the top end, where $L$ is the length of the chain.}
	\label{fig:schema}
%	\vspace{-0.3cm}
\end{figure}

Letting $F(s,t)\geq 0$ be the tension in the chain,
\begin{equation}
	\label{eq:11}
	\|(\rho(s), z(s))\|_2 = \sqrt{\rho'(s)^2 + z'(s)^2} = 1.
\end{equation}
\begin{align}
	\label{eq:x1z}
	(F\rho')' + \mu\rho \omega^2 &= 0, \\
	\label{eq:2}
	(Fz')'  - \mu g &= 0,
\end{align}
where $\Box'$ denotes differentiation with respect to $s$, and $g$ is the gravitational acceleration. Eq.~(\ref{eq:x1z}) and~(\ref{eq:2}) are the equations of motion of the chain, and Eq.~(\ref{eq:11}) shows the inextensibility constraint.

We now derive the dynamics equation for a rotating chain with bottom end fixed. For that, one has to consider the constraint force at the bottom end of the chain to keep it fixed in its position along the \textit{z}-axis. This difference affects the theoretical equations derived from here on, ultimately changing the dimensionality of the problem.
\begin{equation}
	\label{eq:24}
	Fz' = \int_0^s\mu g \; \d \lambda= \mu g s + T.
\end{equation}
Let $T_\textrm{full} \geq 0$ be the constraint force, and $T = T_\textrm{full} z'(0)$ is the component of that force parallel to the \textit{z}-axis. Using the inextensibility constraint~(\ref{eq:11}), we obtain
\begin{equation}
	\label{eq:F}
	F =\frac{\mu g s + T}{z'} =  \frac{\mu g s + T}{\sqrt{1-\rho'^2}}.
\end{equation}
and substituting this into Eq.~(\ref{eq:x1z}) gives the equations which govern the shape function $\rho(s)$
\begin{equation}
	\label{eq:x1}
	\frac{\d {}}{\d s} \left( \frac{\mu g s + T}{\sqrt{1-\rho'^2}}\rho' \right)
	+ \mu\rho\omega^2
	= 0
\end{equation}
which is subjected to the following boundary conditions
\begin{equation}
	\label{eq:1}
	\begin{aligned}
		\rho(L) & = r, \\
		\rho(0) & = 0.
	\end{aligned}
\end{equation}

\subsection{Problem formulation}
\label{sec:probform}

Our paper defines the \emph{configurations} and the \emph{control inputs} of a rotating chain with bottom end fixed as follows:

\begin{definition} \label{def:conf} (Configuration) A configuration of the rotating chain is $q:=(\omega,\rho, T)$, where $\omega \geq 0$ is a rotation speed, $\rho$ is a shape function satisfying the governing equation~(\ref{eq:x1}), and $T$ is the tension at $s=0$ to keep the fixed end in place. In addition, $\rho'(0) \geq 0$ and $z'(0) \geq 0$. The configuration space of the rotating chain, $\calC$, is the set of all such configurations.
\end{definition}

\begin{definition} (Control input) A control input is $(r, \omega, h)$, where $r \geq 0$ is an attachment radius, $\omega \geq 0$ is a rotation speed, and $h \geq 0$ is the vertical distance between the top and bottom end ($z(L)-z(0)$). The control space, $\calV$ is the set of all such inputs.
\end{definition}

%We assume the lower end of the chain is fixed to a flat surface and as such the condition $z'(s) \geq 0$ ensures the chain does not come into contact with that surface.

Given a pair of starting and goal configurations $(q_\mathrm{init}, q_\mathrm{goal})$, the rotating chain manipulation problem involves finding a control trajectory $(0, 1) \rightarrow \calV$ that brings the chain in a stable manner from $q_\mathrm{init}$ to $q_\mathrm{goal}$.

\subsection{Relation to Industrial Settings}
This section will explore how the previously formulated problem can be adapted for specific industrial applications. Both bending and twisting stiffness are assumed to be negligible for a long slender chain. Drill strings belong to a special case of the problem in which the attachment radius-to-height ratio is small (i.e., $r/h$ small), due to limitations arising from the hole dimensions \cite{bailey1960analytical}. In the yarn spinning process, the attachment radius is at the bottom instead of the top end \cite{wang2024experimental}. This can be modeled by simply reversing the direction of gravity in the problem formulation ($g=9.81$ instead of $g=-9.81$). This change reverses the setup such that the rotating chain's top end is fixed, its bottom is allowed to rotate with an attachment radius, and the \textit{z}-axis is flipped. For both cases, the stability analysis method remains consistent with that of the general form. However, the final results for the yarn spinning process are expected to differ because of the different shapes formed.

\section{Forward kinematics of the rotating chain with bottom end fixed}
\label{sec:forward_kinematics}
This section presents the shapes of a rotating chain with bottom end fixed, which are essential for solving the manipulation problem mentioned in Section~\ref{sec:probform}. These shapes are obtained through analysis of the forward kinematics of the rotating chain.

\subsection{Dimensionless shape equation} 
\label{sec:dimless}

Similar to Kolodner, we modify Eq.~(\ref{eq:x1}) into a dimensionless form, making it more suitable for further analyses. The following dimensionless variables are defined
\begin{equation}
	\label{eq:change_of_var}
	u := \frac{\rho'}{\sqrt{1-{\rho'}^2}} \frac{\omega^2}{g} \left(\frac{T}{\mu g} + s\right), \quad
	\bar s := \frac{s\omega^2}{g},
\end{equation}
which can be manipulated to give the expression
\begin{equation}
	\label{eq:rho1}
	\rho' =\frac{u }{\sqrt{{\left(\bar s + \bar T\right)}^2+u^2}}
\end{equation}
and combines with Eq.~(\ref{eq:x1}) to result in
\begin{equation}
	\label{eq:rho2}
	\frac{\d {u}}{\d {\bar s}} + \rho \frac{\omega^2}{g} = 0.
\end{equation}

Through differentiation of Eq.~(\ref{eq:rho2}) with respect to
$\bar s$, one comes to
\[
\frac{\d{}^2  }{\d{\bar s}^2 } u + \rho' = 0,
\]
which is combined with Eq.~(\ref{eq:rho1}) to produce the dimensionless differential equation
\begin{equation}
	\label{eq:new}
	\frac{\d{}^2  }{\d{\bar s}^2 } u(\bar s) +
	\frac{u(\bar s)} {\sqrt{{\left(\bar s + \bar T\right)}^2+{u(\bar s)}^2}} = 0. 
\end{equation}
where 
\begin{equation}
	\label{eq:dimless_T}
	\bar T = \frac{T{\omega}^2}{\mu {g}^2},
\end{equation}
is the dimensionless \textit{z}-axis component of the constraint force at the bottom end of the chain.

From Eq.~(\ref{eq:1}), we can derive the boundary conditions on $u$ as
\begin{equation}
	\label{eq:boundary}
	u'(0) = 0, \quad u'(\bar L) = \bar r,
\end{equation}
where
\begin{equation}
	\label{eq:defa}
	\bar L:=L\omega^2/g, \quad
	\bar r:=-r\omega^2/g. 
\end{equation}
This is the standard form of a Boundary Value Problem (BVP).

\noindent \textbf{Remark} For a certain value of $T$, one can obtain
\begin{equation}
	\label{eq:rm1}
	u(0) = \frac{\rho'(0)}{\sqrt{1-{\rho'(0)}^2}} \bar T,
\end{equation}
and thus,
\begin{equation}
	\label{eq:rm2}
	u''(0) = - c.
\end{equation}
where $c = \rho'(0)$.

% This change in dimensionality of the problem from its free counterpart is interesting to note as the shapes formed is still that of a rotating chain. For the case of the fixed rotating chain, the solution to the shooting method results in not only the shape function, but also the force experienced at the fixed end.

\subsection{Shooting method}
\label{sec:shooting}
To solve the BVP, we employ the simple shooting method. For a given input $(r, \omega, \rho'(0))$, the corresponding $(\bar r, \bar L, c)$ is computed. Next, an initial value of T is guessed to determine $u_{guess}(0)$ in the initial condition $(u(0),u'(0))=(u_{guess}(0),0)$. Then, we integrate this from $\bar s = 0$ to $\bar s=\bar L$ using 10 discretized steps and check if $u'(\bar L)=\bar r$. If they are not equal, refine the guess for $T$ by \emph{e.g.},~ Newton's method and repeat the steps until they are. The $\rho(s)$ from ${u'}_{\rm{last\_iter}}(\bar s)$ obtained in the final iteration gives the solution shape.

Using the inextensibility constraint (\ref{eq:1}), one can also obtain $z(s)$ from $\rho(0)$, $z'(s)\geq 0$, and $z(L)=0$. For each input into the BVP, multiple solutions are obtained.

\subsection{Rotation modes}
\label{sec:modes}
We will now define the rotation mode of a chain with bottom end fixed. The mode is equal to $i$ when the chain crosses its rotation axis (\textit{z}-axis) $i$ times, when viewed along the normal vector of the rotating plane, excluding the bottom end fixed to the rotation axis. In essence, $i$ is equal to the number of zeros in $u'(\bar s)_{\bar s\in(0,\bar L)}$ minus $1$.

\section{Analysis of the configuration space of the rotating chain}
\label{sec:configuration_space}

This section explores the properties of the configuration space of a rotating chain with bottom end fixed, and its stable subspace.

\subsection{Parameterization of the configuration space}
Under the assumption that $(\omega, T, \rho'(0))$ is upper-bounded by $(\omega_{\max}, T_{\max}, 1)$, we can deduce that $(\bar L, \bar T, c)$ are upper-bounded by $(\bar L_{\max}, \bar T_{\max}, 1)$. Using this, we propose that the configuration space can be parameterized by these variables.
\begin{proposition}[and definition]
	\label{prop:A_to_C}
	Define the parameter space $\calA$ by
	$$
	\calA := (0, \bar L_{\max}) \times (0, \bar T_{\max}) \times (0, c_{\max}).
	$$
	% \[
	% \calA := (-a_{\max}, a_{\max}) \times (0, \bar L_{\max}).
	% \]
	where $c_{\max} = 1$. There exists a homeomorphism $f:\calA\to\calC$.
\end{proposition}
Assuming a constant length of the chain, $\bar L$, $\bar T$, and $c$ are functions of $\omega$, $T$, and $\rho'(0)$, respectively. By rewriting Eq.~(\ref{eq:new}) as a dimensionless ODE
\begin{equation}
	\label{eq:X}
	\frac{\d {\vec u}}{\d {\bar s}} = \mathbf{X}(\vec u,\bar s).
\end{equation}
where $\vec u:=(u,u')$, we present a proof for Proposition \ref{prop:A_to_C}.

\begin{proof}
	The mapping $f$ is essentially the shooting method described in Section~\ref{sec:shooting}. Given a $(\bar L, \bar T, c)\in \calA$, we first obtain $\omega$ from $\bar L$ using the relationship $\bar L = L\omega ^2/g$, and $T$ from $\bar T$ using $\bar T = \frac{T{\omega}^2}{\mu {g}^2}$. Next, we integrate the ODE~(\ref{eq:X}) from the initial condition
	\[
	\vec u(0) = ( \frac{c}{\sqrt{1-c^2}} \bar T, 0)
	\]
	until $\bar s=\bar L$ to obtain $u'(\bar s)$ for
	$ \bar s \in (0, \bar L)$.  Finally, we obtain $\rho$ from
	$u'$ using Eq.~\ref{eq:rho2}. 
	
	(1) Surjectivity of $f$. Let $(\omega,\rho, T)\in \calC$. Since $\rho$ verifies~(\ref{eq:x1}), one can perform the change of variables~(\ref{eq:change_of_var}) and obtain $u$ and $u'$. Next, consider $\bar L=L\omega ^2/g$, $\bar T=\frac{T{\omega}^2}{\mu {g}^2}$, and $c=\frac{u(0)}{\sqrt{\bar T^2 + u(0)^2}}$. One has clearly $\bar L\in(0,\bar L_{\max})$, $\bar T\in(0,\bar T_{\max})$, $c\in(0,1)$, and $f(\bar L, \bar T, c) = (\omega,\rho,T)$.
	
	(2) Injectivity of $f$. Assume that there are $(\bar L_1, \bar T_1, c_1) \neq (\bar L_2, \bar T_2, c_2)$ such that $f(\bar L_1, \bar T_1, c_1) = f(\bar L_2, \bar T_2, c_2) = (\omega,\rho,T)$. One has $\bar L_1=\bar L_2=L\omega^2/g$, $\bar T_1=\bar T_2=\frac{T{\omega}^2}{\mu {g}^2}$, and $c_1=c_2=\rho'(0)$, which implies the injectivity.
	
	(3) Continuity of $f$. It is shown in~\cite{pham2017robotic} Appendix~C that the ODE~(\ref{eq:X}) is Lipschitz. It follows that the function $u'(\bar s)$ for $0 \leq \bar s \leq \bar L$ depends continuously on its initial condition, which implies that $\rho(s)$ depends continuously on $\bar T$ and $c$.
		
	(4) Continuity of $f^{-1}$. It can be seen from the injectivity proof that $\bar L$, $\bar T$, and $c$ depend continuously on $\omega$, $T$ and $\rho'(0)$, and $\rho'(0)$ depends in turn continuously on $\rho$.
\end{proof}

By definition of $\calC$, we have proven that the configuration space of the rotating chain with bottom end fixed of constant length is homeomorphic to a three-dimension cube of fixed height ($c\in(0,1)$).

\begin{figure}[htp]
	\centering
	\vspace{-0.3cm}
	\includegraphics[width=0.32\textwidth,trim={0 2.5cm 0 3.5cm},clip]{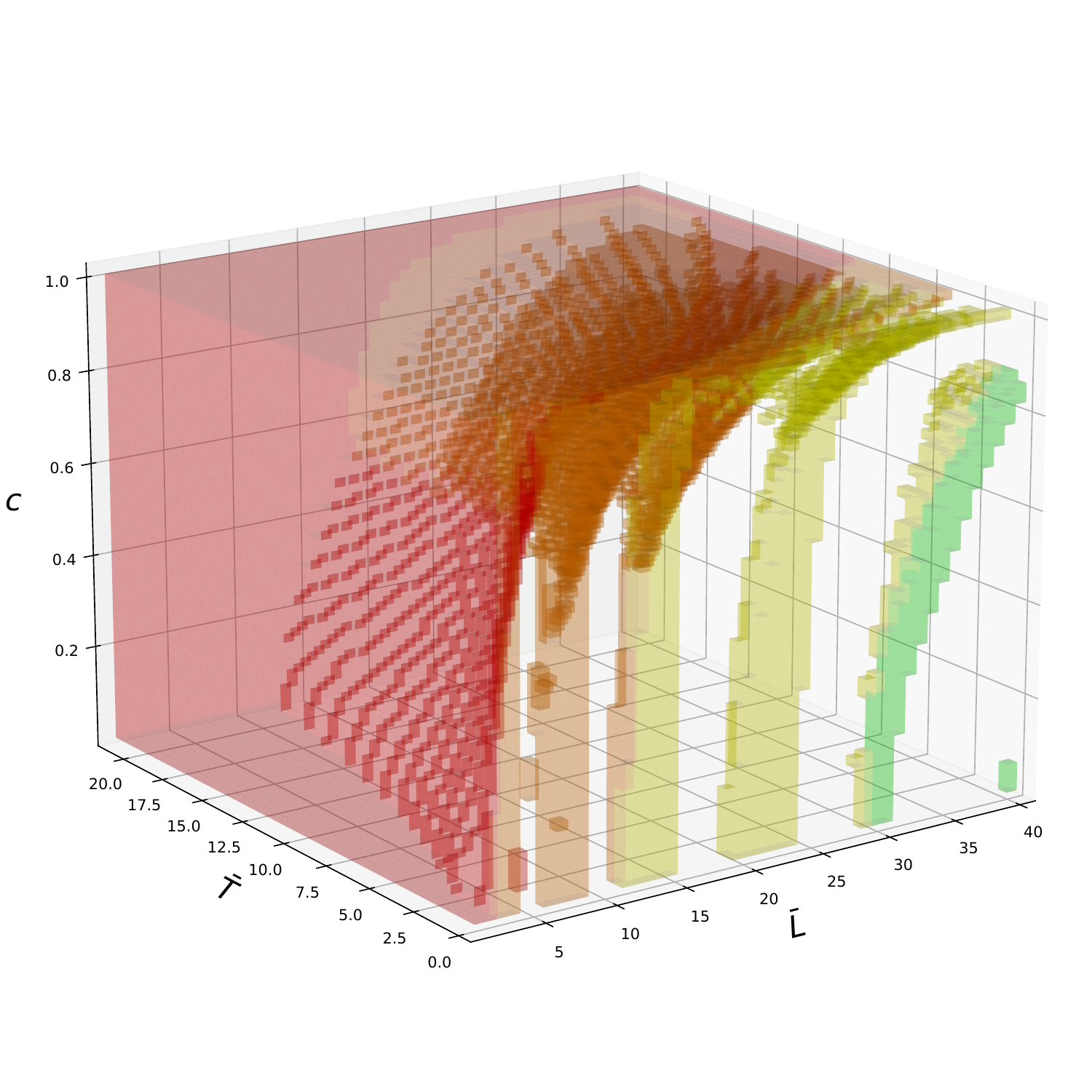}
	\caption{The space of stable configurations of the rotating chain with bottom end fixed. Stability is determined by the largest real part of the eigenvalues of the linearized dynamics of a 10-link lumped-mass model at equilibrium.}
	\label{fig:stabmap}
	\vspace{-0.3cm}
\end{figure}

\subsection{Stability analysis}
\label{sec:stability_analysis}
This section will determine the subspace in which configurations of the rotating chain are stable. First, the chain is approximated with a $N$-link lumped masses model with spring links and its state is represented by $\vec y := [\vec x_1, \dot {\vec x}_1, \dots, \vec x_{N-1}, \dot {\vec x}_{N-1}]$ where $\vec x_i\in \mathbb{R}^{3}$ denotes the position of the $i$-th mass in the rotating frame. The top and bottom free ends are represented with $i = N$ and $i = 0$, respectively. Unlike~\cite{pham2017robotic}, the state of the bottom fixed end ($\vec x_0, \dot {\vec x}_0$) is excluded from stability computations.

%Stability analysis is done by first discretizing the chain into lumped masses with spring links~\cite{pham2017robotic}. $\vec x_i\in \mathbb{R}^{3}$ is the position of the $i$-th mass in the rotating frame $\{\rm O\}$. For a fixed rotating chain, both $\vec x_0$ and $\vec x_N$ are fixed in $\{\rm O\}$. The state of the discretized chain is as follows
%\begin{equation}
%	\label{eq:14}
%	\vec y := [\vec x_0, \dot {\vec x}_0, \dots, \vec x_{N-1}, \dot {\vec x}_{N-1}].
%\end{equation}
Through the use of Newton's laws, the dynamics equation $\dot{\vec y} = \vec f(\vec y)$ including the fictitious Coriolis and centrifugal forces, constraint forces at the links, and aerodynamic forces, is used to derive the Jacobian for stability analysis
\begin{equation}
	\label{eq:jac-stab}
	\vec J(\vec y^{\rm{eq}}):=\frac{\rm d\vec f}{\rm d \vec y}(\vec y^{\rm{eq}}).
\end{equation}
where $\vec y^{\rm{eq}}:= [\vec x^{\rm{eq}}_1, \mathbf{0}, \dots, \vec x^{\rm{eq}}_{N-1}, \mathbf{0}]$ is the chain state in rotational equilibrium. By finding the largest real part of its eigenvalues $\lambda_{\max}:=\max_{i} \rm{Re}(\lambda_{i})$ one can determine if the configuration is asymptotically stable ($\lambda_{\max} \leq 0$), or unstable ($\lambda_{\max} > 0$). Formulating the rotating chain's Jacobian analytically decreased computational time by more than 15 times as compared to computing the numerical Jacobian. Full analytical Jacobian formulation can be found in our Github repository. We find that $N=10$ provides reliable results while keeping computational time reasonable.

\begin{figure}[htp]
	\centering
	\vspace{-0.4cm}
	\includegraphics[width=0.4\textwidth,trim={0 3cm 0 3cm},clip]{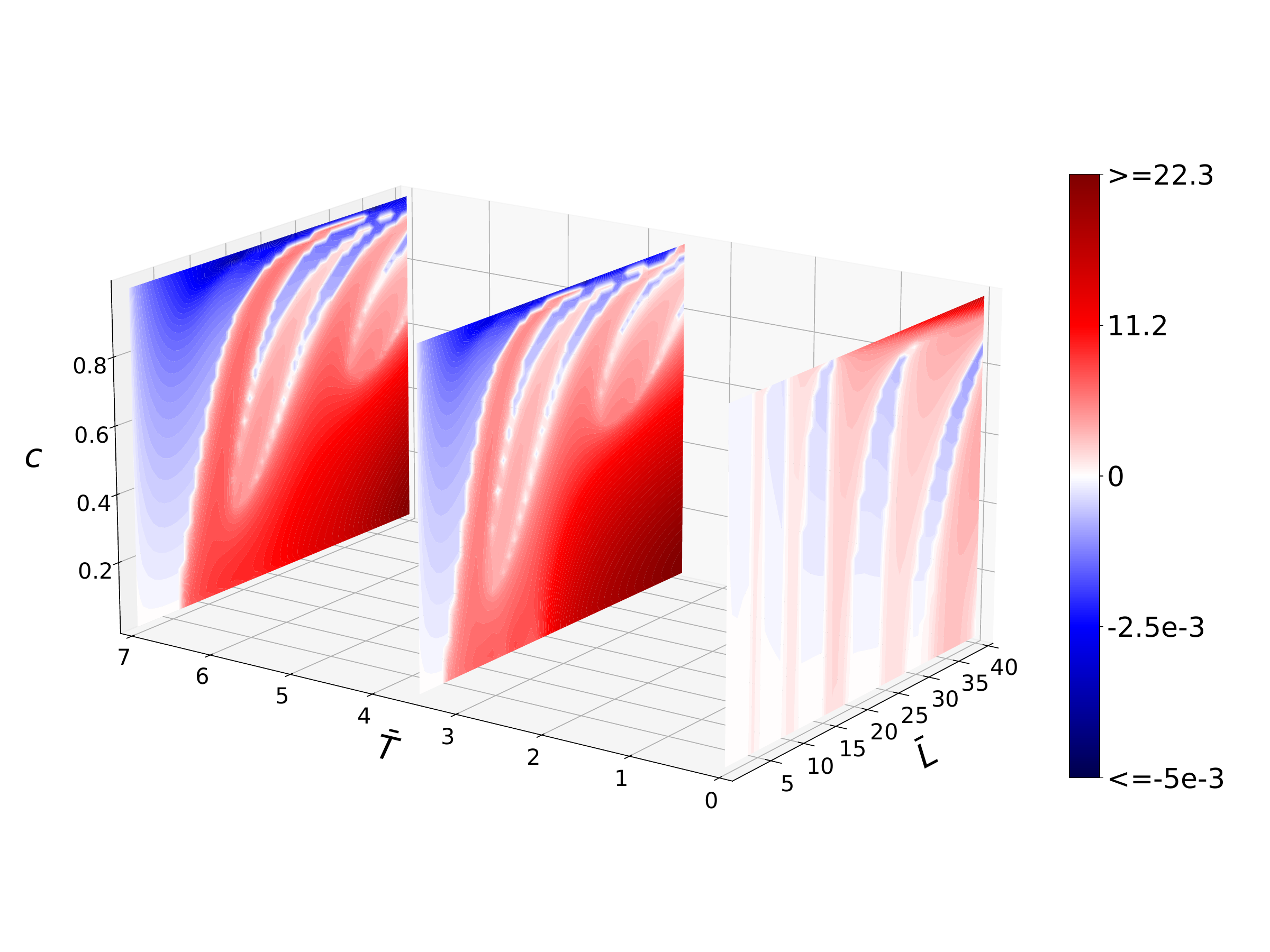}
	\vspace{-0.2cm}
	\caption{A slice of the stability plot. Colors represent the values of $\lambda_{\max}$. Configurations with small $T$ and $c$ values are possible regions where transition between rotation modes can take place.}
	\label{fig:slice_stable}
\end{figure}

Fig.~\ref{fig:slice_stable} shows the stability heat map of three-slices of Fig.~\ref{fig:stabmap} to show the effects of stability when varying $\bar T$. We observe that the stable configuration space is not connected between rotation modes, which could complicate mode transitions. However, at small $\bar T$ and $c$ values, the space between stable rotation modes has $\lambda_{\max}$ values close to 0, suggesting that transitions may be feasible. Indeed, shapes in these region exhibit very small amplitudes (appearing to be at rest) and was experimentally less susceptible to instability.

\section{Manipulation of the rotating chain}
\label{sec:manipulation}
This section outlines how insights into the chain's stability and configuration space facilitate consistent transitions between rotation modes in manipulation planning.

\subsection{Experiment}
\label{sec:experiment}
\begin{figure}[htp]
	\centering
	\textbf{a}\\
	\includegraphics[width=0.35\textwidth,trim={0 2cm 0 3cm},clip]{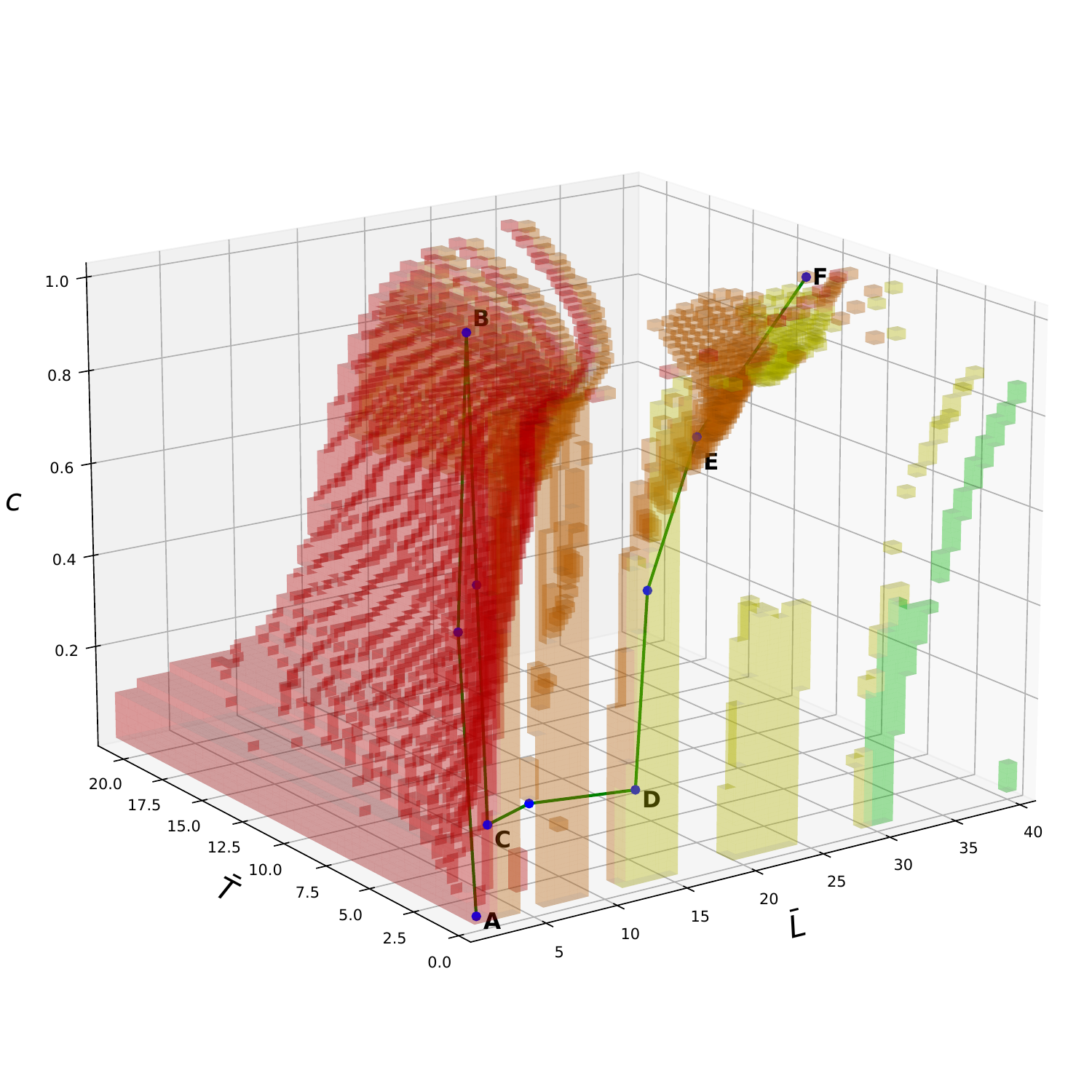}\\
	\textbf{b}\\
	\includegraphics[width=0.45\textwidth,trim={0 0 0 1cm},clip]{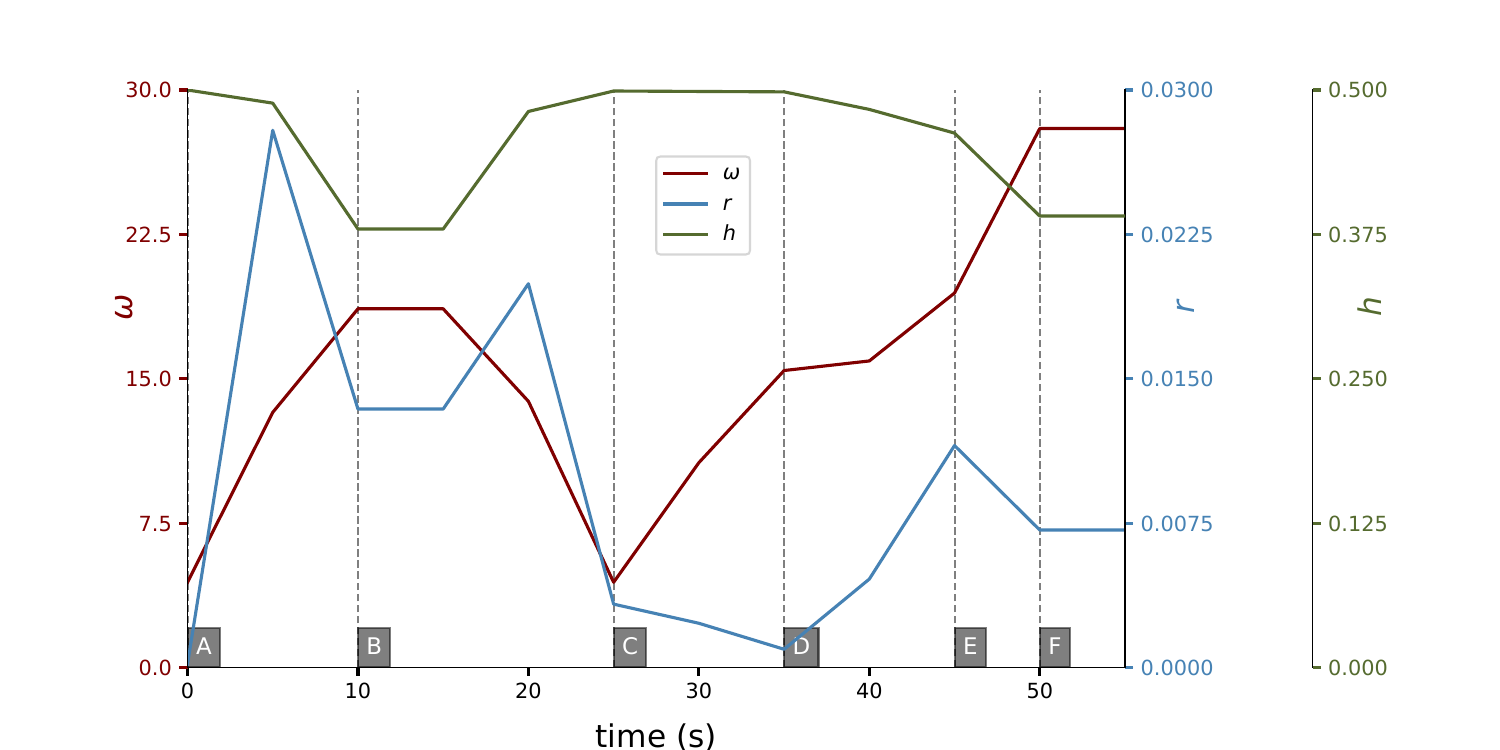}
	\caption{\textbf{a}: The path taken, following the manipulation strategy proposed in Section~\ref{sec:experiment}, through the stable and feasible subspace of the configuration of a rotating chain with bottom end fixed. For an alternative perspective of the 3D space, please refer to the attached video. \textbf{b}: Histories of the control inputs. Each path marker is labeled ([A]-[F]). At the desired first [B] and second [F] rotation mode configurations, control is maintained for $\SI5s$ to clearly observe the shape formed in the physical experiments.}
	\label{fig:exp_control}
	\vspace{-0.3cm}
\end{figure}

% Blue: attachment radius~$r$; brown: angular speed $\omega$; green: height $h$. 

We now propose a manipulation strategy for our physical experiments. By tracing the simplified stable configuration space, the control inputs can be systematically guided to manipulate the chain into a desired shape. Fig.~\ref{fig:exp_control}(a) shows the stable and feasible configuration space of a rotating chain with bottom end fixed. Feasibility is achieved by removing configurations which have control inputs outside of the estimated dynamic and kinematic constraints of the robot arm used. The experiment is carried out on a metallic chain of length $\SI{0.50}{m}$. The top end of the chain is attached to the end-effector of a 6-DOF industrial manipulator (Denso VS-060). Snapshots of the chain at different stages is shown in Fig.~\ref{fig:realshapes}(a), along with their corresponding analytical solutions. The chain was successful in making stable transitions between these configurations. A video of the experiment can be found at \url{https://youtu.be/Hhag8Ea5x30}.

For successful mode transition, planning was conducted such that, for any pair of adjacent nodes ($i$ and $j$), 3 conditions were satisfied:
\begin{enumerate}
	\item The absolute difference in $c$ between adjacent nodes, $K_{i,j} = \left|c_i - c_j\right|$, is such that $0.1 < K_{i,j} < 0.5$. The upper and lower bounds were determined experimentally to prevent unstable rotations and erratic controls, respectively.
	\item The path $\gamma_{i,j}$ between adjacent nodes did not contain regions of instability where $\lambda_{\max} > 1$. That is, $\max_{x \in \gamma_{i,j}} \lambda_{\max}(x) \leq 1$ was satisfied. This condition prevented unstable rotations or the formation of wrong configuration shapes.
	\item All control input $(r, \omega, h)$ lay within the feasible control space of the robot arm.
\end{enumerate}

Since these conditions are placed on the general system, they should theoretically be suitable for all physical and design parameters of the chain and manipulator. To obtain shapes of the first two rotation modes -- [B] and [F] -- from rest [A], the robot arm executed controls dictated by the planned path, where local connections in the feasible and stable subspace between the markers are made through linearly interpolating in the space of controls, as shown in Fig.~\ref{fig:exp_control}(b). Supporting the discussion in Section~\ref{sec:stability_analysis}, mode transition was successfully executed from [C] to [D].
% and mass $\SI{27.7}{g}$

%A classical method (rapidly exploring random tree with path smoothing) was initially employed for planning. However, the planned path infringed the lower bound of condition 1 and the resulting control output was erratic and caused instability. 
Examples of failed transitions include one from A to B without intermediate points which infringed on condition 1, and from B to F in the same manner which breached condition 2, both of which can be found in the video. These experimental results confirm theoretical findings that the defined control space is not homeomorphic to the configuration space of the rotating chain -- a control input can produce multiple configurations. Therefore, successful manipulation requires knowing not just the desired final control input but also a suitable manipulation path to it.

\subsection{Implications for applications}
\label{sec:implappl}
\subsubsection{Drill string operation}
To reduce wear in drill strings, high-frequency oscillations should be avoided~\cite{wada2018longitudinal}. One can achieve this by limiting the rotation speed of the drill string during operation, meaning a smaller $\bar L$ for fixed drill string lengths. Collision of the drill string with the borehole walls~\cite{dong2016review} can be prevented using rotation shapes with a smaller $\max(\rho)$ per unit $h$. Also, rotations should be stable to prevent sudden, uncontrolled drill string collisions. Configurations which satisfy these requirements are stable and tend to be of higher rotation modes. One can reach such configurations by planning in the stable and feasible subspace of the drill string.

\subsubsection{Yarn spinning}
Spinning yarn configurations of higher rotation modes are recommended as they have smaller yarn tension which reduces breakage~\cite{wang2024experimental}. Fewer yarn breakage means higher productivity of the machines in ring spinning. Using our work, one can determine whether and how higher modes can be arrived at through the stability plots. A control trajectory can be obtained from a path planned to the desired rotation mode. In addition, one can conclude that to prevent the yarn from falling back to lower modes during operation, we have to avoid configurations with small $\bar T$ and $c$.

\section{Conclusion}
\label{sec:conclusion}
This paper concludes that the rotating chain with bottom end fixed has a configuration space homeomorphic to a three-dimensional cube. We explored the manipulation problem by addressing the stable transition between various rotation modes. To achieve this, we created a system to analyze the shapes formed by the rotating chain and parameterized its configuration space. From the results obtained, we introduced a manipulation plan for stable and controlled transitions between various rotation modes. Practically, we discussed how our work can contribute to the reduction of drill string fatigue, and the prevention of breakage in yarn spinning.

\subsection*{Limitations and future work}
 Before this work can be implemented in real-world domains, more complex interaction forces -- such as friction, air vortices, and other environmental effects -- must be accounted for. These factors could account for the difficulty in consistently achieving the third and higher rotation modes. To this end, incorporating learning techniques could greatly complement existing theory in modeling real systems. Future work can extend this configuration space simplification to general DLO shape prediction and manipulation planning for other formulations, such as in \cite{rucker2011statics}.

%The formulation of deformable linear objects in \cite{rucker2011statics} presents a promising avenue for the application of this method to shape prediction and manipulation planning in future research.

\addtolength{\textheight}{-0cm}   % This command serves to balance the column lengths

\bibliographystyle{ieeetran}
\bibliography{ref.bib}

\end{document}